\documentclass[11pt,twoside]{article}
\usepackage[T1]{fontenc}
\usepackage{a4wide}
\usepackage{graphics}

\makeatletter

\newcommand{\LyX}{L\kern-.1667em\lower.25em\hbox{Y}\kern-.125emX\@}
\newcommand{\noun}[1]{\textsc{#1}}

\newenvironment{lyxcode}
  {\begin{list}{}{
    \setlength{\rightmargin}{\leftmargin}
    \raggedright
    \setlength{\itemsep}{0pt}
    \setlength{\parsep}{0pt}
    \ttfamily}%
   \item[]}
  {\end{list}}

\makeatother

\begin{document}

\title{Semantic robust parsing for noun extraction \\
from natural language queries}

\author{Afzal Ballim and Vincenzo Pallotta\\
MEDIA group: Laboratoire d'Informatique Théorique (LITH)\\
École Polytechnique Fédérale de Lausanne (EPFL)\\
IN-F Ecublens 1015 Lausanne (Switzerland)\\
Phone:+41-21-693 52 97~~~ Fax:+41-21-693 52 78\\
\texttt{\textbf{\small \{ballim,pallotta\}@di.epfl.ch}}\small }

\date{~}

\maketitle
\begin{abstract}
This paper describes how robust parsing techniques can be fruitful applied for
building a query generation module which is part of a pipelined NLP architecture
aimed at process natural language queries in a restricted domain. We want to
show that semantic robustness represents a key issue in those NLP systems where
it is more likely to have partial and ill-formed utterances due to various factors
(e.g. noisy environments, low quality of speech recognition modules, etc...)
and where it is necessary to succeed, even if partially, in extracting some
meaningful information. 
\end{abstract}

\section{Introduction}

The domain we are concerned with in our case study is the interaction through
speech with information systems. The availability of a large collection of annotated
telephone calls for querying the Swiss phone-book database (i.e the Swiss French
PolyPhone corpus \cite{chollet96:_swiss_frenc_polyp_polyv}) allowed us to experiment
our recent findings in robust text analysis obtained in the context of the Swiss
National Fund research project ROTA (Robust Text Analysis), and in the Swisscom
funded project ISIS (Interaction through Speech with Information Systems). Within
this domain, the goal is to build a valid query to an information system, using
limited world knowledge of the domain in question. Although a task like this
may, at its simplest, be performed quite effectively using heuristic methods
such as keyword spotting, such an approach is brittle, and does not scale up
easily in the case of conducting a dialogue.

\subsection{Problem specification}

In this section we will give an informal specification for the problem of processing
telephone calls for querying a phone-book database.

\subsubsection{Swiss French PolyPhone Database}

These database contains 4293 simulated recordings related to the ``111'' Swisscom
service calls (e.g. ``rubrique 38'' of the calling sheet \cite{chollet96:_swiss_frenc_polyp_polyv}).
Each recording consists of 2 files, one ASCII text file corresponding to the
initial prompt and the information request and one data file containing the
sampled sound version. As far as the address fields are concerned, the data
in the PolyPhone database are unfortunately not tagged and even not consistent.
Prompts and information requests expressed by users have been extracted from
the files an regrouped into a single representation in the following format: 

\begin{verse}
\texttt{\small id:cd1/b00/f0000o06:sid17733~}{\small \par}

\texttt{\small prompt:1~}{\small \par}

\texttt{\small adr1:MOTTAZ~MONIQUE~}{\small \par}

\texttt{\small adr2:rue~du~PRINTEMPS~4~}{\small \par}

\texttt{\small adr3:SAIGNELEGIER~}{\small \par}

\texttt{\small text{[}123{]}:~Bonjour~j'aimerais~un~numéro~de~téléphone~à~Saignelegier
c'est~Mottaz~m~o~deux~ta~z~Monique~rue~du~printemps~numéro~quatre~}{\small \par}

\texttt{\small sample:0.200000:10.820000:88160:42801}{\small \par}
\end{verse}
where currently, the corresponding lines in text file are processed with the
following heuristic:

\begin{description}
\item [id]identifies the original location of the file in the CD-ROM.
\item [prompt]identifies both the type of prompt asking the user for posing the query
(e.g. n. 1 corresponds to ``\emph{Veuillez maintenant faire comme si vous étiez
en ligne avec le 111 pur demander le numéro de téléphone de la personne imaginaire
dont les coordonnées se trouvent ci-dessous:}``).
\item [adr1]corresponds to the \emph{name}
\item [adr2]corresponds to the \emph{address} if line 3 is not empty and \emph{town}
otherwise
\item [adr3]corresponds to the \emph{town} if not empty.
\item [text]corresponds to the text transcription. The number in square brackets is
the total number of chars in the request.
\item [sample]groups the information for the sampled sound version of the request.
\end{description}
This heuristic seems to perform quite well but a more thorough and exhaustive
evaluation still needs to be carried out. The main problem remains in finding
enough information about the \emph{original} data in order to be able to perform
the validation automatically.

\subsubsection{The frame schema}

Concerning the structure in the Swiss Phone-book database, we assumed it is
the same as the one that appears on the web (e.g. http://www.ife.ee.ethz.ch/cgi-bin/etvq/f),
namely (one field per line):

\begin{lyxcode}
{\small Nom~de~famille~/~Firme}{\small \par}

{\small Prénom~/~Autres~informations}{\small \par}

{\small No~de~téléphone}{\small \par}

{\small Rue,~numéro}{\small \par}

{\small NPA,~localité}{\small \par}
\end{lyxcode}
We chose to provide further information which are not available at web level
but which can be used to form the query. The full frame description is given
below\footnote{
Bracketted slots are optional.
}:

\begin{description}
\item [{\footnotesize {[}Caller{]}}]~{\footnotesize \par}

\begin{description}
\item [{\footnotesize Title:}]~{\footnotesize \par}
\item [{\footnotesize Name:}]~{\footnotesize \par}
\item [{\footnotesize Locality:}]~{\footnotesize \par}
\end{description}
\item [{\footnotesize Target\_Identification}]~{\footnotesize \par}

\begin{description}
\item [{\footnotesize Name}]{\footnotesize (default: Person)}{\footnotesize \par}

\begin{description}
\item [{\footnotesize {*}Person}]~{\footnotesize \par}

\begin{description}
\item [{\footnotesize Family~name:}]~{\footnotesize \par}
\item [{\footnotesize {[}Title{]}:}]~{\footnotesize \par}
\item [{\footnotesize {[}First~name{]}:}]~{\footnotesize \par}
\item [{\footnotesize {[}Second~name{]}:}]~{\footnotesize \par}
\item [{\footnotesize {[}Occupation{]}}]~{\footnotesize \par}

\begin{description}
\item [{\footnotesize Description:}]~{\footnotesize \par}
\item [{\footnotesize {[}Class{]}:}]{\footnotesize {[}}\emph{\footnotesize yellow
pages categories}{\footnotesize {]}}{\footnotesize \par}
\end{description}
\end{description}
\item [{\footnotesize {*}Company}]~{\footnotesize \par}

\begin{description}
\item [{\footnotesize Name:}]~{\footnotesize \par}
\item [{\footnotesize {[}Description{]}:}]~{\footnotesize \par}
\item [{\footnotesize {[}Category{]}:}]{\footnotesize {[}}\emph{\footnotesize yellow
pages categories}{\footnotesize {]}}{\footnotesize \par}
\item [{\footnotesize {[}Owner{]}:}]~{\footnotesize \par}
\item [{\footnotesize {[}Contact~person{]}:}]{\footnotesize {[}repres., direction,
secretariat, ...{]}}{\footnotesize \par}
\end{description}
\end{description}
\end{description}
\item [{\footnotesize Target\_Address}]~{\footnotesize \par}

\begin{description}
\item [{\footnotesize {[}Appart~n.{]}:}]~{\footnotesize \par}
\item [{\footnotesize {[}Street~n.{]}:}]~{\footnotesize \par}
\item [{\footnotesize {[}Building{]}:}]~{\footnotesize \par}
\item [{\footnotesize {[}Street~name{]}:}]~{\footnotesize \par}
\item [{\footnotesize {[}Village{]}:}]~{\footnotesize \par}
\item [{\footnotesize {[}NPA{]}:}]~{\footnotesize \par}
\item [{\footnotesize Loc\_type:}]~{\footnotesize \par}
\item [{\footnotesize Locality}]{\footnotesize (at least one of the sub-fields)}{\footnotesize \par}

\begin{description}
\item [{\footnotesize City:}]~{\footnotesize \par}
\item [{\footnotesize ``Environs'':}]~{\footnotesize \par}
\item [{\footnotesize Region:}]~{\footnotesize \par}
\item [{\footnotesize Canton:}]~{\footnotesize \par}
\item [{\footnotesize Telephone~prefix:}]~{\footnotesize \par}
\end{description}
\end{description}
\item [{\footnotesize Request~type}]~{\footnotesize \par}

\begin{description}
\item [{\footnotesize Phone~type:}]{\footnotesize (default: standard) {[}standard,
privé, fax, natel{]}}{\footnotesize \par}
\item [{\footnotesize Request~status:}]{\footnotesize (default: ok) {[}ok, ill-formed,
missing-information, ...{]}}{\footnotesize \par}
\end{description}
\end{description}
One point still remains unclear about the PolyPhone database (as no answers
where found in \cite{andersen97:_swiss_aedvan_vocal_inter_servic,chollet96:_swiss_frenc_polyp_polyv}):
what was the set of annotation used for the transcription of utterances? Several
speech annotations such as ``\texttt{<hesitation>}'' appear in the text. Was
it systematic? Are there other such markers? It is possible to rely on prosodic
informations? In the first phase of the project we simply skipped these informations
but we guess that they could be of great help in disambiguating interpretations
of strict adjacent sequences of names such as in utterances like ``\emph{j'amerais
le numéro de téléphone de Vedo-Moser Brigitte Brignon Baar-Nendaz}''.

\subsection{Query analysis}

The processing of the corpus data is performed at various linguistic levels
performed by modules organized into a pipeline. Each module assumes as input
the output of the preceding module. The main goal of this architecture is to
understand how far it is possible go without using any kind of feedback and
interactions among different linguistic modules.

\subsubsection{Morpho-Syntactic analysis}

At a first stage, morphological and syntactic processing is applied to the output
from the \emph{speech recognizer} module which usually produces a huge word-graph
hypothesis. Low-level processing (morphological analysis and tagging) were performed
by ISSCO (Institute Dalle Molle, University of Geneva) using tools that were
developed in the European Linguistics Engineering project MULTEXT. For syntactic
analysis, ISSCO developed a Feature Unification Grammar based on the ELU formalism
\cite{estival90:_elu_user_manual} (i.e. an extension of PATRII grammars) and
induced by a small sample of the Polyphone data. This grammar was taken by another
of our partners (the Laboratory for Artificial Intelligence of the Swiss Federal
Institute of Technology, Lausanne) and converted into a probabilistic context-free
grammar, which was then initially trained with a sample of 500 entries from
the Polyphone data. The forest of syntactic trees produced by this phase will
be used to achieve two goals:

\begin{enumerate}
\item The n-best analyses are use to disambiguate speech recognizer hypotheses 
\item They served as the input for the robust semantic analysis that we performed,
that had as goal the production of query frames for the information system. 
\end{enumerate}

\subsubsection{Semantic annotations}

While the semantic analysis will in general reduce the degree of ambiguity found
after syntactic analysis, there remains the possibility that it might \emph{increase}
some degree of ambiguity due to the presence of coherent senses of words with
the same syntactic category (e.g., the word ``Geneva'' can refer to either
the canton or the city).

\subsubsection{Semantic robust analysis and frame filling}

The component that deals with such input is generally referred to as a \emph{robust
analyzer}. Although robustness can be considered as being applied at either
a syntactic or semantic level, we believe it is generally at the semantic level
that it is most effective. This robust analysis needs a model of the domain
in which the system operates, and a way of linking this model to the lexicon
used by the other components. It specifies semantic constraints that apply in
the world and which allow us to rule out incoherent requests (for instance).
The degree of detail required of the domain model used by the robust analyzer
depends upon the ultimate task that must be performed --- in our case, furnishing
a query to an information system. Taking the assumption that the information
system being queried is relatively close in form to a relational database, the
goal of the interpretative process is to furnish a query to the information
system that can be viewed in the form of a frame with certain fields completed,
the function of the querying engine being to fill in the empty fields.

One way in which the interface could interact with the querying system would
be to submit such a frame at the end of the analysis process without performing
any coherency checking. The advantage of this method is that the model of the
domain of queries that is required by the interface can be limited. However,
such an approach has two major disadvantages:

\begin{itemize}
\item the result of incorrectly formulated queries may be completely uninterpretable
or erroneous, and the interface system would have no basis for evaluating the
quality of such replies, or how to aid the user in formulating a better one;
\item there might be a number of possible frames that could be submitted for any instance
of a user utterance/query, and this number might be reducible by application
of a model of coherent queries.
\end{itemize}
We will, therefore, presume that queries must be classified by the interface
into three categories:

\begin{enumerate}
\item the query is correct --- the fields of the frame which must be completed contain
semantically valid data. The query may be submitted;
\item incomplete queries --- certain necessary fields cannot be unambiguously filled
in, and so a system-initiative dialogue can be invoked to furbish the necessary
information to create a correct query;
\item incoherent queries --- information in the fields of the frame is not coherent
with the interfaces model of the domain. An error dialogue must be invoked.
\end{enumerate}
The last query category is the most complex, since it requires a domain model
sufficiently rich to decide whether a query is outside of the domain, or inside
the domain but violating certain semantic constraints. In addition, it requires
relatively complex dialogue management as the corrective dialogue may involve
resolution of miscomprehension by either the system or the user.

\section{Computational logic for robust analysis}

What has been considered to be an advantage using \emph{logic-based} programming
languages is the symbol processing capability and the way of abstracting from
the actual implementation of needed data structures. \emph{Definite Clause Grammars}
come to mind when relating Logic Programming and Natural Language Processing.
This is of course one of the best couplings between Computational Linguistics
and Logic to support both (i) the development of linguistic models of Natural
Language (Computational Linguistics) and (ii) the design of real life applications
(Language Engineering). 

The main drawback to this approach is efficiency, but it is not the only one.
In recent years several efforts have been done to improve efficiency of logic
and functional programming languages by means of powerful abstract machines
and optimized compilers. Sometimes, efficiency recovery leads to introduction
of non-logical features in the language and the programmer should be aware of
it in order to exploit it in the development of his or her applications (i.e.
cut in logic programming). 

An important question to ask is: ``how can computational logic contribute to
robust discourse analysis ?''. A partial answer to this question is that currently
logic-based programming languages are able to integrate in an unifying framework
all or most of the techniques necessary for robust text analysis. Furthermore
this can be done in a rigorous ``mathematical'' fashion. In this sense robustness
is related to correctness and provability with respect to the specifications.
A NLP system developed within a logical framework has a predictable behavior
which is useful in order to check the validity of the underlying theories.

\subsection{Left-corner Head-driven Island Parser}

LHIP \cite{ballim94:_lhip,lieske98:rethink_nlp_prolog} is a system which performs
robust analysis of its input, using a grammar defined in an extended form of
the Definite Clause Grammar formalism used for implementation of parsers in
Prolog. The chief modifications to the standard Prolog `grammar rule' format
are of two types: one or more right-hand side (RHS) items may be marked as `heads',
and one or more RHS items may be marked as `ignorable'. 

LHIP employs a different control strategy from that used by Prolog DCGs, in
order to allow it to cope with ungrammatical or unforeseen input. The behavior
of LHIP can best be understood in terms of the complementary notions of \textbf{span}
and \textbf{cover}. A grammar rule is said to produce an island which \textbf{spans}
input terminals \( t_{i} \) to \( t_{i+n} \) if the island starts at the \( i^{th} \)
terminal, and the \( i+n^{th} \) terminal is the terminal immediately to the
right of the last terminal of the island. A rule is said to \textbf{cover} \( m \)
items if \( m \) terminals are consumed in the span of the rule. Thus \( m\leq n \).
If \( m=n \) then the rule has completely covered the span.

As implied here, rules need not cover all of the input in order to succeed.
More specifically, the constraints applied in creating islands are such that
islands do not have to be adjacent, but may be separated by non-covered input.
There are two notions of non-coverage of the input: \textbf{unsanctioned} and
\textbf{sanctioned} non-coverage. The former case arises when the grammar simply
does not account for some terminal. Sanctioned non-coverage means that special
rules, called ``\emph{ignore}'' rules, have been applied so that by ignoring
parts of the input the islands are adjacent. Those parts of the input that have
been \emph{ignored} are considered to have been consumed. These \emph{ignore}
rules can be invoked individually or as a class. It is this latter capability
which distinguishes \emph{ignore} rules from regular rules, as they are functionally
equivalent otherwise, but mainly serve as a notational aid for the grammar writer.

Strict adjacency between RHS clauses can be specified in the grammar. It is
possible to define global and local thresholds for the proportion of the spanned
input that must be covered by rules; in this way, the user of an LHIP grammar
can exercise quite fine control over the required accuracy and completeness
of the analysis.

A chart is kept of successes and failures of rules, both to improve efficiency
and provide a means of identifying unattached constituents. In addition, feedback
is given to the grammar writer on the degree to which the grammar is able to
cope with the given input; in a context of grammar development, this may serve
as notification of areas to which the coverage of the grammar might next be
extended. Extensions of Prolog DCG grammars in LHIP permit: 

\begin{enumerate}
\item nominating certain RHS clauses as heads; 
\item marking some RHS clauses as being optional; 
\item invocation of \emph{ignore} rules; 
\item imposing adjacency constraints between two RHS clauses; 
\item setting a local threshold level in a rule for the fraction of spanned input
that must be covered. 
\end{enumerate}
A threshold defines the minimum fraction of terminals covered by the rule in
relation to the terminals spanned by the rule in order for the rule to succeed.
For instance, if a rule spans terminals \( t_{i} \) to \( t_{i+n} \) covering
\( j \) terminals in that span, then the rule can only succeed if \( j/n\geq T \).

The following is an example of a LHIP rule. At first sight this rule appears
left recursive. However, the sub-rule ``\texttt{conjunction(Conj)}'' is marked
as a head and therefore is evaluated before either of ``\texttt{s(Sl)}'' or
``\texttt{s(Sr)}''. Presuming that the conjunction-rule does not end up invoking
(directly or indirectly) the s-rule, then the s-rule is not left-recursive. 

\begin{lyxcode}
s(conjunct(Conj,Sl,Sr)) \~{}\~{}>~\\
~~~~~~~~~~~s(Sl)~\\
~~~~~~~~~~{*}conjunction(Conj),~~\\
~~~~~~~~~~~s(Sr).
\end{lyxcode}
LHIP provides a number of ways of applying a grammar to input. The simplest
allows one to enumerate the possible analyses of the input with the grammar.
The order in which the results are produced will reflect the lexical ordering
of the rules as they are converted by LHIP. With the threshold level set to
0, all analyses possible with the grammar by deletion of input terminals can
be generated. By setting the threshold to 1, only those partial analyses that
have no unaccounted for terminals within their spans can succeed. Thus, supposing
a suitable grammar, for the sentence \emph{John saw Mary and Mark saw them\/{}}
there would be analyses corresponding to the sentence itself, as well as \emph{John
saw Mary}, \emph{John saw Mark}, \emph{John saw them}, \emph{Mary saw them},
\emph{Mary and Mark saw them,} etc. By setting the threshold to 1, only those
partial analyses that have no unaccounted for terminals within their spans can
succeed. Hence, \emph{Mark saw them\/{}} would receive a valid analysis, as
would \emph{Mary and Mark saw them\/{}}, provided that the grammar contains
a rule for conjoined NPs; \emph{John saw them\/{}}, on the other hand, would
not. As this example illustrates, a partial analysis of this kind may not in
fact correspond to a true sub-parse of the input (since \emph{Mary and Mark\/{}}
was not a conjoined subject in the original). Some care must therefore be taken
in interpreting results. 

This rule illustrates a number of features: \emph{negation}, and \emph{optional
forms}. The rule will only succeed if (with respect to the area of input in
which it might occur) there is a noun with no determiner. In addition, there
can be optional adjectives before the noun. 

\begin{lyxcode}
np(propernoun(N,Mods))~\~{}\~{}>~~\\
~~~~~~~~~~~~~~~~\~{}~determiner(\_),~~\\
~~~~~~~~~~~~~~~~~(?~adjectives(Mods)~?),~\\
~~~~~~~~~~~~~~~~{*}~noun(N).
\end{lyxcode}
This rule illustrates the use of disjunction and embedded Prolog code. It should
be noted that within the scope of a disjunction or negation, a head is local
to the disjunct or negation.

\begin{lyxcode}
noun(X)~\~{}\~{}>~~\\
~~~~~(~{*}~@pussy,~(?~@cat~?);~{*}~@cat),~~\\
~~~~~\{X=cat\}.
\end{lyxcode}
This rule illustrates a typical use of adjacency, to specify compound nouns.
Adjacency is not restricted such a use however, but may generally be used anywhere.

\begin{lyxcode}
noun(missionary\_camp)~\~{}\~{}>~@missionary~:~@camp.
\end{lyxcode}
A number of tools are provided for producing analyses of input by the grammar
with certain constraints. For example, to find the set of analyses that provide
maximal coverage over the input, to find the subset of the maximal coverage
set that have minimum spans, and to find the find analyses that have maximal
thresholds. In addition, other tools can be used to search the chart for constituents
that have been found but are not attached to any complete analysis. The conversion
of the grammar into Prolog code means that the user of the system can easily
develop analysis tools that apply different constraints, using the given tools
as building blocks.

\section{Implementation of the semantic module}

In our approach we try to integrate the above principles in our system in order
to effectively compute hypotheses for the frame filling task. This can be done
by building a lattice of \emph{frame filling hypotheses} and possibly selecting
the best one. Hypotheses are typically sequences of proper names. The lattice
of hypotheses is generated by means of LHIP \emph{discourse grammar}. This type
of grammar is used to extract \emph{names chunks} and assemble them into the
hypothesized frame structure.

\subsection{Tree-paths representation}

Parse trees obtained from the previous module are encoded into a path representation
which allows us to easily specify constraints over the tree structure. A \emph{path-sentence}
is a list of \emph{path-words} which in turn are compound terms of the type
\texttt{terminal(word,~path)} where \emph{word} is a constant term and \emph{path}
is a list of arc identifiers that is compound terms \texttt{'cat'(\#number\_of\_nodes,
\#node,~\#identifier)} uniquely identifying an arc in the parse tree. The functor
\texttt{'cat'} is a category name and its arguments are integer positive numbers.
For instance the representation of the parse tree:

{\par\centering \resizebox*{0.2\textwidth}{0.15\textheight}{\includegraphics{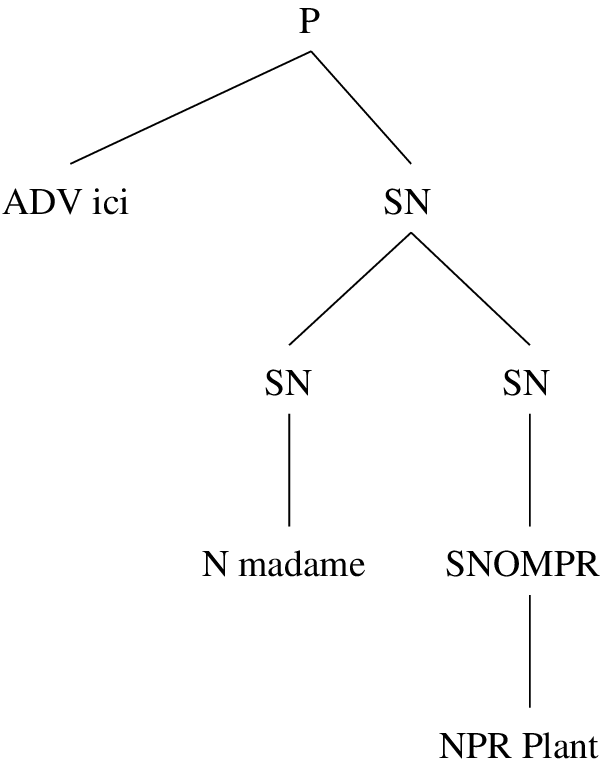}} \par}

is given by:

\texttt{\scriptsize {[}terminal(ici,{[}'ADV'(1,1,14),'P'(2,1,12),'P'(2,1,11)),}{\scriptsize \par}

\texttt{\scriptsize terminal(madame,{[}'N'(1,1,19),'SN'(1,1,17),'SN'(2,1,16),'P'(2,2,15),'P'(2,1,11)),}{\scriptsize \par}

\texttt{\scriptsize terminal('Plant',{[}'NPR'(1,1,24),'SNOMPR'(1,1,22),'SN'(1,1,21),'SN'(2,2,20),'P'(2,2,15),'P'(2,1,11){]}.}{\scriptsize \par}

Using this representation it is possible to define a grouping operator (e.g.
\texttt{group/2}) which given a sequence of adjacent names finds the subsequence
of words having the least common ancestor which is closer than the least common
ancestor (e.g. \texttt{lca/2}) of the given sequence. These two operators are
very useful for imposing structural knowledge constraints and they are straightforwardly
defined as PROLOG programs by:

\begin{lyxcode}
{\footnotesize lca({[}terminal(\_,W){]},W).}~\\
{\footnotesize lca({[}terminal(\_,W)|R{]},P)~:-~}~\\
{\footnotesize ~~~~~~~~lca(R,P1),~}~\\
{\footnotesize ~~~~~~~~prefix\_path(P1,P),~}~\\
{\footnotesize ~~~~~~~~prefix\_path(W,P),!.}~\\
{\footnotesize ~~}{\footnotesize \par}

{\footnotesize group({[}{]},{[}{]}).~}~\\
{\footnotesize group(L,X)~:-~}~\\
{\footnotesize ~~~~~~~~lca(L,P),~}~\\
{\footnotesize ~~~~~~~~proper\_sublist(L,X),~length(X,N),~N>1,~}~\\
{\footnotesize ~~~~~~~~lca(X,P1),~}~\\
{\footnotesize ~~~~~~~~proper\_sublist(P1,P).}~\\
{\footnotesize ~}{\footnotesize \par}

{\footnotesize prefix\_path(A,A).~}~\\
{\footnotesize prefix\_path({[}\_|B{]},C)~:-~}~\\
{\footnotesize ~~~~~~~~prefix\_path(B,C).}{\footnotesize \par}
\end{lyxcode}

\subsection{Discourse markers}

Discourse segments allow us to model dialog by a set of pragmatic concepts (dialogue
acts) representing what the user is expected to utter (for example initiation
of a dialogue: \emph{init}, expression of gratitude: \emph{thank}, and demand
for information: \emph{request}, etc.) and in that way are useful for reducing
the syntactic and semantic ambiguity. These are domain-dependent and must be
defined for a given corpus. For their definition, we intend to follow the experiments
done in the context of Verbmobil (see for example \cite{jekat95:_dialog, kompe94:_prosod}).
In our specific case identifying special words serving both as separators among
logical subparts of the same sentence and as introducers of semantic constituents
allows us to search for name sequences to fill a particular slot only in interesting
part of the sentence. One of the most important separator is the \emph{announcement-query
separator}. The LHIP clauses defining this separator can be one or more word
covering rule like for instance:

\begin{lyxcode}
{\small ann\_query\_separator~\#1.0~\~{}\~{}>~}~\\
{\small ~~~~~~~~@terminal('téléphone',\_).~}{\small \par}

{\small ann\_query\_separator~\#1.0~\~{}\~{}>~}~\\
{\small ~~~~~~~~(~@terminal('numéro',\_):~}~\\
{\small ~~~~~~~~~~@terminal('de',\_):~}~\\
{\small ~~~~~~~~~~(?~@terminal('téléphone',\_)~?)).}{\small \par}
\end{lyxcode}
As an example of semantic constituents introducers we propose here the 

\begin{lyxcode}
{\small street\_intro({[}T,Prep,Det{]},1)~\#1.0~\~{}\~{}>~}~\\
{\small ~~~~~~~~{*}~street\_type(T),~}~\\
{\small ~~~~~~~~preposition(Prep),~}~\\
{\small ~~~~~~~~determiner(Det).~}{\small \par}
\end{lyxcode}
which make use of some word knowledge about street types coming from an external
thesaurus like:

\begin{lyxcode}
{\small street\_type(terminal(X,P))~\~{}\~{}>~}~\\
{\small ~~~~~~~~@terminal(X,P),~}~\\
{\small ~~~~~~~~\{thesaurus(street,W),member(X,W)\}.}{\small \par}
\end{lyxcode}

\subsection{Generation of hypotheses}

The generation of hypotheses for filling the frame is performed by: composing
weighted rules, assembling chunks and filtering possible hypotheses.

\subsubsection{Weighted rules}

The main assumption on which probabilistic approach to NLP is based, is that
language is considered as being a random phenomenon with its own probability
distribution function: \emph{coverage} is often translated as \emph{expectation}
in a probabilistic sense. Changing perspective and considering language just
as an \emph{uncertain} and \emph{imprecise} phenomenon and understanding as
a \emph{perception} process, it is naturally to think of \emph{fuzzy} models
of language (see \cite{lee69:_note} and \cite{asveld96:_towar_robus_fuzzif}).
Recently, fuzzy reasoning has been partially integrated into a CLP paradigm
(see \cite{riezler96:_quant_const_logic_progr_weigh_gramm_applic}) in order
to deal with so called \emph{soft constraints} in weighted \emph{constraint
logic grammars}. We tried to get some inspiration from the above proposal for
integrating fuzzy logic and parsing to compute weights to assign to each frame
filling hypotheses. Each LHIP rule returns a confidence factor together with
the sequence of names. The confidence factor for a rule can be either assigned
statically (e.g. to pre-terminal rules) or they can be computed composing recursively
the confidence factors of sub-constituents. Confidence factors are combined
choosing the minimum among confidences of each sub-constituents. It is possible
that there is no enough information for filling a slot. In this case the grammar
should provide a mean to provide an empty constituent when all possible hypothesis
rules have failed. This is possible using negation and epsilon-rules in LHIP
as showed in the following rules for dealing with street names.

\begin{lyxcode}
{\small found\_street\_name(L,Conf)~\#1.0~\~{}\~{}>~}~\\
{\small ~~~~~~~~{*}~street\_intro(Intro,Conf),~}~\\
{\small ~~~~~~~~name\_list(X),~}~\\
{\small ~~~~~~~~\{append(Intro,X,L)\}.}{\small \par}

{\small found\_street\_name(X,0.3)~\~{}\~{}>~}~\\
{\small ~~~~~~~~{*}~name\_list(X).}{\small \par}

{\small hyp\_street\_name(Street,Conf)~\~{}\~{}>~}~\\
{\small ~~~~~~~~{*}~found\_street\_name(Street,Conf).}{\small \par}

{\small hyp\_street\_name({[}{]},1)~\~{}\~{}>~}~\\
{\small ~~~~~~~~\~{}found\_street\_name(\_,\_),~}~\\
{\small ~~~~~~~~lhip\_true.}{\small \par}
\end{lyxcode}
where \texttt{\small name\_list(X)} accounts for sequence of adjacent proper
names and \texttt{\small lhip\_true} corresponds to the empty sequence. 

Observe that in this particular case there is no need to select the minimum
confidence factor from the sub-constituents of the rule \texttt{\small found\_street\_name}
since we have only \texttt{\small street\_intro(Intro,Conf)} which propagates
its confidence factor.

\subsubsection{Chunk assembling}

The highest level constituent is represented by the whole frame structure which
simply specifies the possible orders of chunks relative to slot hypotheses.
A rule for a possible frame hypothesis is:

\begin{lyxcode}
{\footnotesize frame(Caller\_title,~Caller\_name,~}~\\
{\footnotesize ~~~~~~Target\_title,~Target\_name,~}~\\
{\footnotesize ~~~~~~Street\_name,~Street\_number,~}~\\
{\footnotesize ~~~~~~Locality,~Weight)~}~\\
{\footnotesize ~~~~~~~~~~~~~\~{}\~{}>~hyp\_caller(Caller\_title,Caller\_name,C1),~}~\\
{\footnotesize ~~~~~~~~~~~~~~~~~{*}~ann\_query\_separator,~}~\\
{\footnotesize ~~~~~~~~~~~~~~~~~hyp\_target(Target\_title,Target\_name,C2),~}~\\
{\footnotesize ~~~~~~~~~~~~~~~~~{*}~location\_intro,~}~\\
{\footnotesize ~~~~~~~~~~~~~~~~~hyp\_street\_name(Street\_name,C3),~}~\\
{\footnotesize ~~~~~~~~~~~~~~~~~hyp\_street\_number(Street\_number,C4));~}~\\
{\footnotesize ~~~~~~~~~~~~~~~~~hyp\_locality\_name(Locality,C5),~}~\\
{\footnotesize ~~~~~~~~~~~~~~}{\footnotesize \par}

{\footnotesize ~~~~~~~~~~~~~~~~~\{minlist({[}C1,C2,C3,C4,C5{]},Weight)\}.~}{\footnotesize \par}
\end{lyxcode}
In this rule we specify a possible order of chunks interleaved by separators
and introducers. The computation of global weight may be more complex than the
above rule which uses simply the minimum of each hypothesis confidence values.
In this case we did not provide any structural constraint (e.g. preferring names
chunks belonging to the minimal common sub-tree or those having the longest
sequence of name belonging to the same sub-tree).

\subsubsection{Filtering and query generation}

The obtained frame hypotheses can be further filtered by both using structural
knowledge (e.g. constraints over the tree-path representation) and word knowledge.
In order to combine the information extracted from the previous analysis step
into the final query representation which can be directly mapped into the database
query language we will make use of a frame structure in which slots represent
information units or attributes in the database. A simple notion of context
can be useful to fill by default those slots for which we have no explicit information.
For doing this type of \emph{hierarchical reasoning} we exploit the meta-programming
capabilities of logic programming and we used a meta-interpreter which allows
multiple inheritance among logical theories \cite{BroTur95}. More precisely
we made use of the special \emph{retraction} operator ``\( \prec  \)'' for
composing logic programs which allows us to easily model the concept of inheritance
in hierarchical reasoning. The expression \( P\prec Q \), where \( P \) and
\( Q \) are meta-variables used to denote arbitrary logic programs, means that
the resulting logic programs contains all the definition of \( P \) except
those that are also defined in \( Q \). 

The definition of the \emph{isa} operator is obtained combining the retraction
operator with the union operator (e.g. \( \cup  \)) that simply make the physical
union of two logic programs, by 
\[
P\; isa\; Q=P\cup (Q\prec P).\]

As an example for the above definition we provide some default definitions which
have been used to represent part of the world knowledge in our domain. The \emph{rules}
theory contains rules for inferring the locality or the locality type when they
are not explicitly mentioned in the query.

\begin{description}
\item [rules:]~
\end{description}
\begin{lyxcode}
locality(City)~:-

~~~~~~~~caller\_prefix(X),

~~~~~~~~prefix(X,City).~\\
~

loc\_type(Type)~:-

~~~~~~~~locality(City),~

~~~~~~~~gis(City,Type).
\end{lyxcode}
where \texttt{prefix/2} and \texttt{gis/2} are world knowledge bases (i.e. a
collection of facts grouped in a theory called \emph{kb}) and \texttt{caller\_prefix/1}
can be easily provided from the answer system. 

If some information is missing then the system tries to provide some default
additional information to complete the query. The following theory contains
definition for some mandatory slots which need to be filled in case of incomplete
queries, like for instance in the theory \emph{query\_defaults}:

\begin{description}
\item [query\_defaults:]~
\end{description}
\begin{lyxcode}
identification(person).

phone\_type(standard).

loc\_type(city).
\end{lyxcode}
Finally starting from an incomplete query which does not account for the required
information we can use deduction to generate the query completion like for instance
asking for:

\begin{lyxcode}
?-~demo((query~\emph{isa}~query\_default)~\( \cup  \)~rules~\( \cup  \)~kb),~loc\_type(X)).
\end{lyxcode}

\section{Conclusions}

From a very superficial observation of the human language understanding process,
it appears clear that no deep competence of the underlying structure of the
spoken language is required in order to be able to process acceptably distorted
utterances. On the other hand, the more experienced is the speaker, the more
probable is a successful understanding of that distorted input. How can this
kind of fault-tolerant behavior be reproduced in an artificial system by means
of computational techniques? Several answers have been proposed to this question
and many systems implemented so far, but no one of them is capable of dealing
with robustness as a whole. 

As examples of robust approaches applied to dialogue systems we cite here two
systems which are based on similar principles.

In the \noun{dialogos} human-machine telephone system (see \cite{albesano97:_dialog})
the robust behavior of the \emph{dialogue management} module is based both on
a contextual knowledge base of pragmatic-based expectations and the dialogue
history. The system identifies discrepancies between expectations and the actual
user behavior and in that case it tries to rebuild the dialogue consistency.
Since both the domain of discourse and the user's goals (e.g. railway timetable
inquiry) are clear, it is assumed the systems and the users cooperate in achieving
reciprocal understanding. Under this underlying assumption the system pro-actively
asks for the query parameters and it is able to account for those spontaneously
proposed by the user. 

In the \texttt{\noun{syslid}} project (see \cite{boros96:_proces_spoken_dialog_system_exper_syslid_projec})
where a robust parser constitutes the \emph{linguistic component} (LC) of the
\emph{query-answering dialogue system} . An utterance is analyzed while at the
same time its semantical representation is constructed. This semantical representation
is further analyzed by the \emph{dialogue control} \emph{module} (DC) which
then builds the database query. Starting from a \emph{word graph} generated
by the speech recognizer module\emph{,} the robust parser will produce a search
path into the word graph. If no complete path can be found, the robust component
of the parser, which is an island based chart parser (see \cite{hanrieder95:_robus_parsin_spoken_dialog_using}),
will select the maximal consistent partial results. In this case the parsing
process is also guided by a \emph{lexical semantic knowledge base} component
that helps the parse in solving structural ambiguities.

We can conclude that robustness in dialogue is crucial when the artificial system
takes part in the interaction since inability or low performance in processing
utterances will cause unacceptable degradation of the overall system. As pointed
out in \cite{Allen96:_robus_system_natur_spoken_dialog} it is better to have
a dialogue system that tries to guess a specific interpretation in case of ambiguity
rather than ask the user for a clarification. If this first commitment results
later to be a mistake a robust behavior will be able to interpret subsequent
corrections as repair procedures to be issued in order to get the intended interpretation.

\bibliographystyle{plain}

\end{document}